# From Qualitative to Quantitative Probabilistic Networks

**Silja Renooij** and **Linda C. van der Gaag**
Institute of Information and Computing Sciences, Utrecht University
P.O. Box 80.089, 3508 TB Utrecht, The Netherlands
{silja,linda}@cs.uu.nl

## Abstract

Quantification is well known to be a major obstacle in the construction of a probabilistic network, especially when relying on human experts for this purpose. The construction of a qualitative probabilistic network has been proposed as an initial step in a network's quantification, since the qualitative network can be used to gain preliminary insight in the projected network's reasoning behaviour. We extend on this idea and present a new type of network in which both signs and numbers are specified; we further present an associated algorithm for probabilistic inference. Building upon these semi-qualitative networks, a probabilistic network can be quantified and studied in a stepwise manner. As a result, modelling inadequacies can be detected and amended at an early stage in the quantification process.

## 1 Introduction

The formalism of probabilistic networks [1] is generally considered an intuitively appealing and powerful formalism for capturing knowledge from a complex problem domain, along with its uncertainties. The graphical structure of the network encodes variables and the probabilistic relationships between them. With associated conditional probabilities it captures the strengths of these relationships. The construction of a probabilistic network typically sets out with the configuration of the graphical structure, before the task of assessing the required probabilities is commenced. Experience shows that, although it may take considerable time, the configuration of a network's graphical structure is quite doable. It is the assessment of the typically large number of probabilities required that is the most daunting, especially when domain experts are the only source of probabilistic information available [2]. Research on facilitating probability assessment for probabilistic networks has thus far focused on elicitation methods that are tailored to the elicitation of a large number of probabilities [3].

Recently, another approach has been advocated [4] that builds upon the use of *qualitative probabilistic networks*. A qualitative probabilistic network in essence is a qualitative abstraction of a probabilistic network [5]. It has the same graphical structure as its quantitative counterpart, but instead of quantifying the probabilistic relationships between the variables by conditional probabilities it summarises these by qualitative signs. For inference with a qualitative probabilistic network, an efficient algorithm is available, based on the idea of propagating signs [6]. This algorithm provides for studying the reasoning behaviour of a probabilistic network in the making prior to its quantification.

We elaborate on the idea of using a qualitative network to facilitate quantification and introduce a methodology that provides for stepwise quantifying a probabilistic network. When the graphical structure of a network in the making is considered robust, a domain expert is asked to associate signs with it to arrive at a qualitative network. Specifying signs is known to require considerably less effort from domain experts than specifying numbers [6]. The construction of the qualitative network will therefore take relatively little time. The qualitative network is then used to perform an initial study of the reasoning behaviour of the probabilistic network under construction.

When quantifying a probabilistic network with the help of domain experts, quantification efforts typically are focused on small parts of the network at a time. As, in each step, conditional probability distributions become available for the variables in the network, we replace the appropriate signs with this numerical information, which results in a network in which both signs and probabilities are specified. Before proceeding to the next part of the network for quantification, the reasoning behaviour of the intermediate network is studied. Modelling inadequacies in the graphical structure can thus be detected and amended at an early stage in the quantification process. This process of quantifying small parts of the network and studying reasoning behaviour is repeated until the network is fully quantified.

To support the methodology of stepwise quantification outlined above, we introduce the formalism of *semi-

*qualitative probabilistic networks* to capture networks in which both signs and probabilities are employed to describe the probabilistic relationships between variables. In addition, we present an efficient algorithm for inference with a semi-qualitative network.

The paper is organised as follows. In Section 2, we provide some preliminaries from the field of probabilistic networks and their qualitative abstractions. In Section 3, we introduce the formalism of semi-qualitative probabilistic networks; the inference algorithm is presented in Section 4. In Section 5, we discuss some complexity issues concerning inference in semi-qualitative networks. Section 6 illustrates our quantification methodology with an example network. The paper is rounded off with some conclusions and directions for further research in Section 7.

## 2 Preliminaries

A *probabilistic network* is a concise representation of a joint probability distribution on a set of statistical variables [1]. It encodes, in an acyclic directed graph $G = (V(G), A(G))$, the variables concerned by means of the set of nodes $V(G)$ and the probabilistic relationships between them by means of a set of arcs $A(G)$. Associated with each node $A$ is a set of conditional probability distributions $\Pr(A \mid \pi(A))$ describing the relationship of this node with its (immediate) predecessors $\pi(A)$ in the digraph. Figure 1(a) shows an example of a simple probabilistic network with three binary-valued nodes.

A probabilistic network defines a unique joint probability distribution on its nodes from which probabilities of interest can be computed. For this purpose, various algorithms are available [1, 7]. These algorithms have an exponential computational complexity in general. For networks with relatively sparse digraphs, as in fact are found in most real-life applications, the algorithms tend to have a runtime complexity that is polynomial in the number of nodes.

*Qualitative probabilistic networks* in essence are qualitative abstractions of probabilistic networks and thus bear a strong resemblance to their quantitative counterparts. A qualitative probabilistic network also comprises an acyclic digraph modelling variables and the probabilistic relationships between them. Instead of conditional probability distributions, however, a qualitative probabilistic network associates with its digraph *qualitative influences* and *qualitative synergies* [5].

A qualitative influence between two nodes expresses how the values of one node influence the probabilities of the values of the other node. Such an influence is summarised by a *sign*. A *positive qualitative influence*, for example, of a node $A$ on its (immediate) successor $B$, denoted $S^+(A, B)$, expresses that observing higher values for $A$ makes higher values for $B$ more likely, regardless of any other direct influence on $B$; for binary-valued nodes $A$ and $B$ with $a > \bar{a}$ and $b > \bar{b}$, this means that

$$\Pr(b \mid ax) - \Pr(b \mid \bar{a}x) \geq 0$$

for any combination of values $x$ for the set $\pi(B) \setminus \{A\}$ of (immediate) predecessors of $B$ other than $A$. A *negative qualitative influence*, denoted by $S^-$, and a *zero qualitative influence*, denoted by $S^0$, are defined analogously, replacing $\geq$ in the above formula by $\leq$ and $=$, respectively. If the influence of node $A$ on node $B$ is not monotonic or if it is unknown, we say that it is *ambiguous*, denoted $S^?(A, B)$. Figure 1(b) shows our example network abstracted to a qualitative probabilistic network; the signs of the qualitative influences are indicated over the arcs.

The set of influences of a qualitative probabilistic network exhibits various convenient properties [5]. The property of *symmetry* guarantees that, if the network includes the influence $S^\delta(A, B)$, $\delta \in \{+, -, 0, ?\}$, then it also includes $S^\delta(B, A)$. The property of *transitivity* asserts that qualitative influences along a chain that specifies at most one incoming arc for each node, combine into a single net influence whose sign is given by the $\otimes$-operator from Table 1. The property of *composition* asserts that multiple qualitative influences between two nodes along parallel chains combine into a single net influence whose sign is given by the $\oplus$-operator.

From the $\oplus$-operator in Table 1, we have that combining parallel qualitative influences with the $\oplus$-operator may yield an ambiguous sign. Such an ambiguity, in fact, results whenever influences with opposite signs are combined. We say that the *trade-off* that is reflected by the conflicting influences cannot be *resolved*. Note that, in contrast with the $\oplus$-operator, the $\otimes$-operator cannot introduce ambiguities upon combining signs of influences along chains. It will cause ambiguous signs to be spread throughout the network once they have arisen, though.

In addition to influences, a qualitative probabilistic network includes product synergies that express how the value of one node influences the probabilities of the values of another node in view of a given value for a third node [8]. The sign of the product synergy serves to capture the sign of the *intercausal influence* it induces between the predecessors of an observed node. The intercausal influence is a qualitative influence in essence and behaves accordingly.

For reasoning with a qualitative probabilistic network, an efficient algorithm is available [6]; this algorithm is sum-marised in pseudocode in Figure 2. The basic idea of the algorithm is to trace the effect of observing a node's value on the other nodes in the network by message-passing between neighbouring nodes. For each node, a *sign* is determined, indicating the direction of change in the node's probabilities occasioned by the new observation given all previous ones. Initially, all node signs equal '0'. For the newly observed node, an appropriate sign is entered, that is, either a '+' for the observed value *true* or a '−' for the value *false*. The node updates its sign and subsequently sends a message to each (induced) neighbour that is not independent of the observed node. The sign of this message is the ⊗-product of the node's (new) sign and the sign of the influence it traverses. This process is repeated throughout the network, building on the properties of symmetry, transitivity, and composition of influences. Since each node can change its sign at most twice (once from '0' to '+', '−' or '?', and then only to '?'), the process visits each node at most twice and therefore halts in polynomial time.

| $\otimes$ | + | − | 0 | ? | $\oplus$ | + | − | 0 | ? |
|---|---|---|---|---|---|---|---|---|---|
| + | + | − | 0 | ? | + | + | ? | + | ? |
| − | − | + | 0 | ? | − | ? | − | − | ? |
| 0 | 0 | 0 | 0 | 0 | 0 | + | − | 0 | ? |
| ? | ? | ? | 0 | ? | ? | ? | ? | ? | ? |

Table 1: The $\otimes$- and $\oplus$-operators for combining signs.

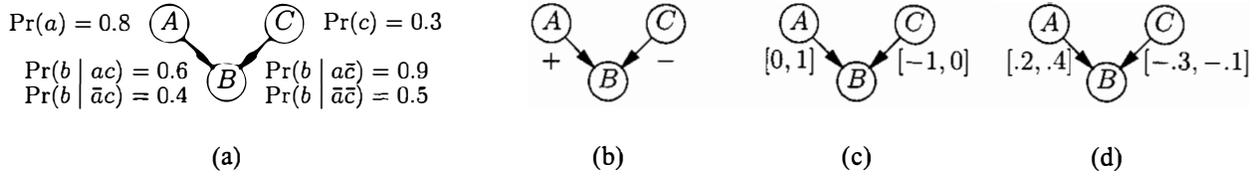

(a)  (b)  (c)  (d)

Figure 1: A probabilistic network fragment (a), its abstraction into a qualitative probabilistic network (b), the interval network equivalent to the qualitative network (c), and the more informed interval network (d).

**procedure** PropagateSign(*trail,from,to,messagesign*):
sign[*to*] ← sign[*to*] ⊕ *messagesign*;
*trail* ← *trail* ∪ {*to*};
**for** each active neighbour $V_i$ of *to*
**do** *linksign* ← sign of (induced) influence between *to* and $V_i$;
*messagesign* ← sign[*to*] ⊗ *linksign*;
**if** $V_i \notin trail$ **and** sign[$V_i$] ≠ sign[$V_i$] ⊕ *messagesign*
**then** PropagateSign(*trail,to,$V_i$,messagesign*).

Figure 2: The sign-propagation algorithm.

## 3 Semi-qualitative networks

A *semi-qualitative probabilistic network* comprises an acyclic digraph modelling statistical variables and the relationships between them, just like a probabilistic network and its qualitative counterpart. Associated with this digraph are conditional probability distributions and signs so as to satisfy the following property: for each node $A$, *either* a set of distributions $\Pr(A \mid \pi(A))$ is specified, *or* each incoming arc $C \to A$, $C \in \pi(A)$, for $A$ has associated a qualitative influence $S^\delta(C, A)$, $\delta \in \{+, -, 0, ?\}$.

Associated with a semi-qualitative probabilistic network, we construct an *interval network* that will be exploited upon inference. In this interval network, each arc $A \to B$ has associated an interval influence, denoted $F^{[p,q]}(A, B)$, where the interval $[p, q] \subseteq [-1, 1]$ has the following meaning: $F^{[p,q]}(A, B)$ if and only if

$$\Pr(b \mid ax) - \Pr(b \mid \bar{a}x) \in [p, q]$$

for all combinations of values $x$ for the set $\pi(B) \setminus \{A\}$ of predecessors of $B$ other than $A$. An interval influence and its associated interval $[p, q]$ will be called positive if $p \geq 0$, negative if $q \leq 0$, zero if $p = q = 0$, and ambiguous otherwise.

To construct an associated interval network, we observe that the signs of a semi-qualitative probabilistic network can be readily interpreted as intervals. For a node $B$ and its predecessor $A$, we have, for example, that

$$S^+(A, B) \iff F^{[0,1]}(A, B)$$

Similarly, a negative influence is an influence with the interval $[-1, 0]$, a zero influence has the interval $[0, 0]$, and an ambiguous influence has $[-1, 1]$. In the sequel, these four intervals will be referred to as the *unit intervals*. The network from Figure 1(c) is the interval network associated with the qualitative network from Figure 1(b). Using the above translation of signs into intervals, the operators from Table 1 can be taken to be operators on intervals as is reflected in Table 2.

We further observe that the conditional probability distributions $\Pr(A \mid \pi(A))$ specified for a node $A$ can be used to compute the interval influences to be associated with $A$'s incoming arcs. As an example, we construct the interval network for the probabilistic network from Figure 1(a). For the arc $A \to B$ we find that

$$\Pr(b \mid ac) - \Pr(b \mid \bar{a}c) = 0.6 - 0.4 = 0.2$$
$$\Pr(b \mid a\bar{c}) - \Pr(b \mid \bar{a}\bar{c}) = 0.9 - 0.5 = 0.4$$

The interval influence of $A$ on $B$ thus is $F^{[0.2, 0.4]}(A, B)$.

| ⊗ | [0,1] | [−1,0] | [0,0] | [−1,1] |
|---|---|---|---|---|
| [0,1] | [0,1] | [−1,0] | [0,0] | [−1,1] |
| [−1,0] | [−1,0] | [0,1] | [0,0] | [−1,1] |
| [0,0] | [0,0] | [0,0] | [0,0] | [0,0] |
| [−1,1] | [−1,1] | [−1,1] | [0,0] | [−1,1] |

| ⊕ | [0,1] | [−1,0] | [0,0] | [−1,1] |
|---|---|---|---|---|
| [0,1] | [0,1] | [−1,1] | [0,1] | [−1,1] |
| [−1,0] | [−1,1] | [−1,0] | [−1,0] | [−1,1] |
| [0,0] | [0,1] | [−1,0] | [0,0] | [−1,1] |
| [−1,1] | [−1,1] | [−1,1] | [−1,1] | [−1,1] |

Table 2: The ⊗- and ⊕-operators for combining intervals.

Note that the interval indicates that the qualitative influence of $A$ on $B$ indeed is positive. For the interval influence of $C$ on $B$ we find $F^{[-0.3,-0.1]}(C, B)$. The resulting interval network is shown in Figure 1(d). Obviously, the interval network contains less information than the fully quantified probabilistic network. In Section 5 we will elaborate on this loss of information.

We are not the first to propose the use of intervals in reasoning with uncertainty; we refer to, for example, [9, 10] for an overview. Intervals have been used to indicate the uncertainty about or imprecision of the actual value of a probability. In our interval network, however, we use intervals to indicate a *range of differences* in probability. As the semantics of our intervals diverge from the semantics that have been proposed before, we feel that the available interval-propagation algorithms are unsuitable for probabilistic inference in our interval networks.

## 4 Inference in a semi-qualitative network

For reasoning with a semi-qualitative network, we introduce an algorithm that operates upon the associated interval network. Our algorithm is closely related to the sign-propagation algorithm discussed in Section 2, and is based on the idea of propagating intervals over arcs. We recall that the sign-propagation algorithm builds on the properties of symmetry, transitivity and composition of qualitative influences. We revisit these properties with respect to interval influences before presenting our propagation algorithm.

### 4.1 Transitivity

To address the effect of transitively combining interval influences, we consider the network fragment from Figure 3. The fragment includes the nodes $A$, $B$ and $C$, with two influences between them; $X$ denotes the set of predecessors of $B$ other than $A$, and $Y$ is the set of all predecessors of $C$ other than $B$. For the indirect influence of node $A$ on node $C$, we observe that

$\Pr(c \mid axy) - \Pr(c \mid \bar{a}xy) =$
$(\Pr(c \mid by) - \Pr(c \mid \bar{b}y)) \cdot (\Pr(b \mid ax) - \Pr(b \mid \bar{a}x))$

for any combination of values $x$ for the set of nodes $X$ and any combination of values $y$ for $Y$. From this expression, we have, for example, that the largest difference in probabilities yielded by the net influence of $A$ on $C$ equals the largest difference in probabilities obtained from the products of the differences in probabilities yielded by the influences from which it is composed. We conclude that

$$F^{[p,q]}(A, B) \land F^{[r,s]}(B, C) \Rightarrow F^{[p,q]\otimes_i [r,s]}(A, C)$$

where $\otimes_i$ denotes the interval multiplication defined in Table 3. The above observations are readily generalised to any chain between two nodes that specifies at most one incoming arc per node.

| $\otimes_i$ | $[r, s]$ |
|---|---|
| $[p, q]$ | $[\min\{p \cdot r, p \cdot s, q \cdot r, q \cdot s\}, \max\{p \cdot r, p \cdot s, q \cdot r, q \cdot s\}]$ |

Table 3: The $\otimes_i$-operator for interval multiplication.

### 4.2 Parallel composition

For combining multiple interval influences between two nodes along parallel chains, we consider the network fragment from Figure 4. The fragment includes the two parallel chains $A \to C$ and $A \to B \to C$ between the nodes $A$ and $C$; $X$ denotes the set of all predecessors of $B$ other than $A$, and $Y$ is the set of predecessors of $C$ other than $A$ and $B$. For the net influence of node $A$ on node $C$ along the two parallel chains, we find that

$\Pr(c \mid axy) - \Pr(c \mid \bar{a}xy) =$
$\quad (\Pr(c \mid aby) - \Pr(c \mid a\bar{b}y)) \cdot \Pr(b \mid ax) + \Pr(c \mid a\bar{b}y)$
$\quad -(\Pr(c \mid \bar{a}by) - \Pr(c \mid \bar{a}\bar{b}y)) \cdot \Pr(b \mid \bar{a}x) - \Pr(c \mid \bar{a}\bar{b}y)$

for any combination of values $x$ for the set of nodes $X$ and any combination of values $y$ for the set $Y$. Now suppose that all influences in the network fragment under consideration are positive with intervals $[p, q]$, $[r', s']$ and $[r'', s'']$, respectively. We thus have $F^{[p,q]}(A, C)$, $F^{[r',s']}(A, B)$, and $F^{[r'',s'']}(B, C)$, with $p, q, r', s', r'', s'' \geq 0$. Further suppose that the interval influences $F^{[r',s']}(A, B)$ and $F^{[r'',s'']}(B, C)$ combine into the indirect influence $F^{[r,s]}(A, C)$. To determine the interval for the net influence of node $A$ on node $C$, we observe that $\Pr(b \mid ax) - \Pr(b \mid \bar{a}x) \in [r', s']$. The lower- and upper-bounds of the interval $[r', s']$ are attained, for example, for $\Pr(b \mid ax) = r'$ and $\Pr(b \mid \bar{a}x) = 0$, and $\Pr(b \mid ax) = 1$ and $\Pr(b \mid \bar{a}x) = 1 - s'$, respectively. For the first situation, we find that

$\Pr(c \mid axy) - \Pr(c \mid \bar{a}xy) \geq$
$(\Pr(c \mid aby) - \Pr(c \mid a\bar{b}y)) \cdot r' + \Pr(c \mid a\bar{b}y) - \Pr(c \mid \bar{a}\bar{b}y)$

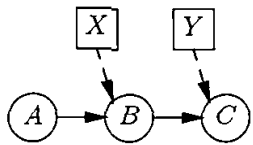

Figure 3: A fragment of a network.

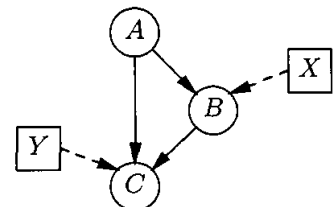

Figure 4: Another network fragment.

The observation that $\Pr(c \mid ab y) - \Pr(c \mid a\bar{b}y) \in [r'', s'']$ and $\Pr(c \mid a\bar{b}y) - \Pr(c \mid \bar{a}\bar{b}y) \in [p, q]$, now gives

$$\Pr(c \mid axy) - \Pr(c \mid \bar{a}xy) \geq r' \cdot r'' + p = p + r$$

for any combination of values $xy$ for the set of nodes $X \cup Y$. Similarly, we find that

$$\Pr(c \mid axy) - \Pr(c \mid \bar{a}xy) \leq s' \cdot s'' + q = q + s$$

We conclude that

$$F^{[p,q]}(A, C) \wedge F^{[r,s]}(A, C) \Rightarrow F^{[p,q] \oplus_i [r,s]}(A, C)$$

where $\oplus_i$ denotes the interval addition operator given in Table 4. The above observations again are readily generalised. For example, if the direct influence of node $A$ on node $C$ in Figure 4 is negative, that is, $F^{[p,q]}(A, C)$ with $p, q \leq 0$, then the lower-bound for the difference $\Pr(c \mid axy) - \Pr(c \mid \bar{a}xy)$ is attained using the smallest value for the direct influence of node $A$ on node $C$; this is again the lower-bound $p$, which is the largest negative value. Whether the net influence of node $A$ on node $C$ now becomes positive, negative, or ambiguous, depends on the actual values of $p, q, r$ and $s$. Note that as the bounds represent differences in probability, they can be no smaller than $-1$ and cannot exceed $+1$.

| $\oplus_i$ | $[r, s]$ |
|---|---|
| $[p, q]$ | $[p + r, q + s] \cap [-1, 1]$ |

Table 4: The $\oplus_i$-operator for interval addition.

### 4.3 Symmetry

We recall from Section 2 that qualitative influences are symmetric. The same observation holds for the 'sign' of an interval influence; for example, if the interval of an influence is known to be positive, then so is the interval of the reverse influence. Interval influences, however, are not symmetric with respect to the interval itself, that is, $F^{[p,q]}(A, B)$ does not necessarily imply $F^{[p,q]}(B, A)$. To provide for propagating intervals against the direction of arcs upon inference, we propose to explicitly specify intervals for reverse influences. We use a *default* unit interval for this purpose. For an arc $A \to B$ with the positive interval influence $F^{[p,q]}(A, B)$, we specify $F^{[0,1]}(B, A)$; for a negative influence $F^{[p,q]}(A, B)$, we specify $F^{[-1,0]}(B, A)$, and so on.

As numerical information becomes available for the nodes in a semi-qualitative probabilistic network, the (default) intervals specified for reverse influences can be tightened. We consider, as an example, a root node $A$ with a single direct successor $B$. Suppose that for node $A$ the probability distribution $\Pr(a) = x$ has been specified. Further suppose that the arc $A \to B$ has associated a positive interval influence. The possible effect of observing node $B$ on node $A$ is then restricted to the interval $[0, \max\{x, 1 - x\}]$. The possible effect of node $B$ on node $A$ is restricted to the interval $[-\max\{x, 1 - x\}, 0]$ if the influence of node $A$ on node $B$ has associated a negative interval. Moreover, for arbitrary nodes $A$ and $B$ with an arc $A \to B$ between them, we have that if we are able to compute the values $\Pr(a \mid bx) - \Pr(a \mid \bar{b}x)$ for all relevant nodes $X$, then we can also determine the interval that is to be associated with the reverse influence of $B$ on $A$. These values can be readily computed by applying Bayes' theorem, if the probability distributions for both nodes $A$ and $B$ are available. After quantifying a node $A$ therefore, we can tighten not just the intervals associated with its incoming arcs, but also the intervals of the reverse influences associated with those outgoing arcs $A \to B$ for which node $B$ has been quantified as well.

### 4.4 The interval-propagation algorithm

In the foregoing, we have shown that interval influences exhibit the properties of transitivity and parallel composition. We have further specified a means of determining reverse interval influences, thereby providing for the property of symmetry. Building upon these properties, we can now use the sign-propagation algorithm for the purpose of propagating intervals, by simply replacing the $\otimes$- and $\oplus$-operators for combining signs by the $\otimes_i$- and $\oplus_i$-operators for combining intervals. With the resulting algorithm, for each node an interval is determined, indicating the upper- and lower-bounds of the change in the node's probabilities occasioned by the new observation, given all previous observations. Initially, all node intervals equal $[0, 0]$. For the newly observed node, an interval $[\alpha, \beta]$ is entered to indicate the strength of the observation. For example, the observation $A = a$ for a node $A$ with $\Pr(a) = x$ is entered as $[1 - x, 1 - x]$. If we have no knowledge about the prior probability of the observed node, then this ignorance is reflected by entering the unit interval $[0, 1]$ for a positive observation and $[-1, 0]$ for a negative observation. Note that we also allow entering imprecise knowledge of the observed node's prior probability.

## 5 Loss of information and complexity

In the previous section, we have detailed our interval-propagation algorithm for probabilistic inference with a semi-qualitative network. Here we address the computational complexity of the basic algorithm and focus on two types of information loss from which it suffers.

### 5.1 Coping with information loss due to abstraction

In Section 3, we have demonstrated that constructing an interval network from a semi-qualitative probabilistic network may result in some loss of information. This loss of information arises from the abstraction of differences in probability to intervals. Since we have defined our interval-propagation algorithm to operate upon an interval network, the algorithm cannot fully exploit all probabilistic information that is available. As a result, it is possible that trade-offs that are modelled in the semi-qualitative network cannot be resolved, even though the available probabilistic information would allow us to do so.

A closely related problem has been addressed by C.-L. Liu and M.P. Wellman [11]. They propose to reason with a probabilistic network in a qualitative way, thereby exploiting the efficiency of sign-propagation, and to revert to the full quantification only when a trade-off leads to an ambiguous result. They describe two methods for resolving the trade-off. The first method amounts to marginalising over the nodes along the conflicting chains that give rise to the trade-off. Nodes are removed in a stepwise manner, using arc reversal and node reduction [12], until the trade-off is resolved or no more nodes are available for removal. For the former successors of the removed nodes, the marginalisation results in updated (conditional) probabilities, which are again abstracted into qualitative signs for further processing. The second method proposed by Liu and Wellman is to estimate bounds on the net influence along the chains that give rise to the trade-off. These bounds are then used to compute the sign of the net influence.

With our methodology of stepwise quantification of a probabilistic network, typically small coherent parts of a network are quantified at a time. Whenever a cluster of nodes involved in a trade-off has been quantified and the interval-propagation algorithm results in an ambiguous interval, then one of the methods from Liu and Wellman can be used to attempt to locally resolve the trade-off. Note that both methods provide us with sufficient information to establish an interval for the net influence, which can then again be used in interval-propagation.

### 5.2 Coping with information loss due to propagation

When discussing the interval-propagation algorithm in Section 4, we have argued that ignorance about the strength of an observation can be expressed by entering the unit interval $[0,1]$ or $[-1,0]$, depending on the sign of the observation. A major drawback of using an interval including a zero as bound, however, is that upon propagation all computed node intervals end up including a zero, which in turn may result in ambiguous intervals. Instead of using the intervals $[0,1]$ and $[-1,0]$, therefore, we propose to enter the intervals $[1,1]$ or $[-1,-1]$, respectively. After propagation, each node interval then describes the *maximum* possible effect of the observation, without taking the actual strength into account. The minimum possible effect is a zero effect, that is, no change. If knowledge about the strength $[\alpha,\beta]$ of the observation becomes available at a later stage, then the node intervals resulting from the propagation can be multiplied with this interval using the $\otimes_i$-operator.

### 5.3 Computational complexity

The interval-propagation algorithm presented in Section 4 closely resembles the sign-propagation algorithm for probabilistic inference in a qualitative network. We recall that with the sign-propagation algorithm a node can change sign at most twice. As a node does not have to pass on any messages when its sign has not changed, the algorithm halts after a number of steps that is polynomial in the number of nodes of the network. A node interval, however, can change as often as the node is visited. The interval-propagation algorithm as a consequence has a worst-case computational complexity that is exponential in the number of nodes. It therefore is not as efficient as its look-alike sign-propagation algorithm.

To ameliorate the problem of an exponential computational complexity, we propose to add a parameter $m$ to the interval-propagation algorithm that serves to limit the number of times a node's interval can be changed. When the $m$th change to the interval occurs, it is set to the unit interval corresponding to the 'sign' of the current interval. For example, if a node's interval is positive after having been visited $m-1$ times, it is set to $[0,1]$ upon the $m$th visit. If, on subsequent visits of the node, the 'sign' of the interval does not change, we do not change the interval at all; if the sign does change, however, then the interval is changed to the appropriate unit interval. Note that once a node has associated a unit interval, it can change at most one more time. Also note that thus restricting the number of changes to a node's interval does not lead to incorrect results upon inference. It just causes results to be less informative.

## 6 An example

In our methodology for stepwise quantifying a probabilistic network, we take its graphical structure for a point of departure. A domain expert is asked to associate signs with the arcs of the structure to arrive at a qualitative network that allows for an initial study of the reasoning behaviour of the probabilistic network under construction. In each following step, quantification efforts are focused on small coherent parts of the network. As a result, conditional probability distributions become available for small clusters of related nodes. This probabilistic information is used to build a semi-qualitative network, from which an interval network is constructed. The reasoning behaviour of the semi-qualitative network can then be studied through interval propagation in its associated interval network. This process is repeated until we have arrived at a fully quantified probabilistic network. In this section we present an example of the use of semi-qualitative probabilistic networks as sketched in the above.

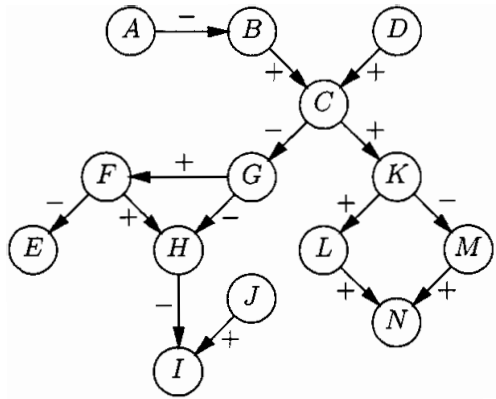

Figure 5: Initial (semi-) qualitative probabilistic network.

The first step after configuring the graphical structure of our example network is to elicit signs from the domain experts. Suppose that the network from Figure 5 is the resulting qualitative network. The reasoning behaviour of this network can be studied using the sign-propagation algorithm. For example, the effect of entering a '+' for node $B$ on all other nodes is shown in the following table, where the ambiguous signs for the nodes $H$, $I$ and $N$ reflect the trade-offs modelled for $H$ and $N$, respectively:

| nodes | node sign |
|---|---|
| $B, C, E, K, L$ | + |
| $A, F, G, M$ | − |
| $D, J$ | 0 |
| $H, I, N$ | ? |

Suppose that in the next step, the nodes $A$ and $B$ are quantified by the domain experts. They indicate that the prior probabilities for node $A$ are $\Pr(a) = 0.4$ and $\Pr(\bar{a}) = 0.6$; the conditional probabilities for node $B$ are $\Pr(b \mid a) = 0.2$, $\Pr(\bar{b} \mid a) = 0.8$, $\Pr(b \mid \bar{a}) = 0.4$, and $\Pr(\bar{b} \mid \bar{a}) = 0.6$. We substitute this probabilistic information in the qualitative network from Figure 5, thereby obtaining a semi-qualitative network. For the associated interval network, we now compute non-unit intervals for the arc $A \to B$. The interval influence of $A$ on $B$ is determined from the conditional probabilities specified for node $B$: we find that $F^{[-0.2,-0.2]}(A, B)$. As nodes $A$ and $B$ are both quantified, we can use Bayes' theorem to determine that $\Pr(a \mid b) = 0.82$ and $\Pr(a \mid \bar{b}) = 0.92$. As a result we find $F^{[-0.1,-0.1]}(B, A)$ for the reverse influence.

To determine the effect of a positive observation for node $B$ on all other nodes in the network, we use the interval-propagation algorithm with the value $[1, 1]$ as suggested in Section 5. The results, indicating the *maximum* possible effect of $B$'s observation on the other nodes, are as follows:

| nodes | node interval |
|---|---|
| $B$ | $[1, 1]$ |
| $C, E, K, L$ | $[0, 1]$ |
| $A$ | $[-0.1, -0.1]$ |
| $F, G, M$ | $[-1, 0]$ |
| $D, J$ | $[0, 0]$ |
| $H, I, N$ | $[-1, 1]$ |

As the available probabilistic information provides for computing the prior probabilities for node $B$: $\Pr(b) = 0.22$ and $\Pr(\bar{b}) = 0.78$, we know that a positive observation occasions a change in $B$'s probabilities of 0.78. The effect of the observation on the probabilities of the other nodes can now be determined by multiplying the above intervals by $[0.78, 0.78]$.

Now suppose that the probabilities for node $D$ happen to be known from the literature: $\Pr(d) = 0.3$ and $\Pr(\bar{d}) = 0.7$. As node $D$ is a root node with a single direct successor, we can tighten the interval $[0, 1]$ of the reverse influence associated with the arc $D \to C$. We find that the upper-bound of the interval is $\max\{0.3, 0.7\}$ and we thus find $F^{[0,0.7]}(C, D)$.

Now suppose that conditional probabilities are provided for the nodes $C$, $K$, $L$, $M$ and $N$. The intervals computed from the newly available probabilistic information for the influences to be associated with the various arcs, are shown in Figure 6; the intervals for the reverse influences are not specified in the figure. Once again we determine the influence of a positive observation for node $B$ on the other nodes in the network, using the interval-propagation algorithm. We find the following results:

| nodes | node interval |
|---|---|
| $B$ | $[1, 1]$ |
| $L$ | $[0.39, 0.50]$ |
| $K$ | $[0.56, 0.72]$ |
| $C$ | $[0.7, 0.9]$ |
| $N$ | $[0.02, 0.32]$ |
| $E$ | $[0, 0.9]$ |
| $A$ | $[-0.1, -0.1]$ |
| $M$ | $[-0.36, -0.28]$ |
| $F, G$ | $[-0.9, 0]$ |
| $D, J$ | $[0, 0]$ |
| $H, I$ | $[-0.9, 0.9]$ |

Note that the trade-off for node $N$ has now been resolved. The intervals $[-0.9, 0.9]$ for the nodes $H$ and $I$, however, still indicate an ambiguity.

Suppose that after quantification of the nodes $F$, $G$ and $H$, interval-propagation still results in an ambiguous influence of node $B$ on node $H$. We now apply Liu and Wellman's method as suggested in Section 5, to attempt to resolve the trade-off involved. Suppose that with the available numerical information, node $F$ is removed by marginalisation, and that by doing so the trade-off is resolved: the net influence of node $G$ on node $H$ over the parallel composition of the two influences has become negative. The new node interval for node $H$ is now the product of the node interval of node $G$ and the interval associated with the computed net influence of $G$ on $H$. As the node interval for node $H$ has now changed, node $H$ sends a new message to node $I$.

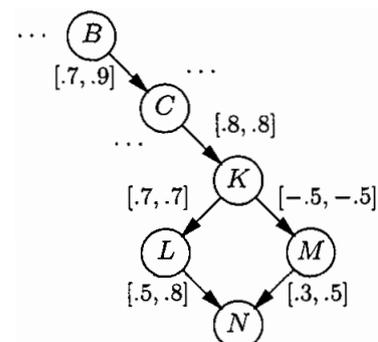

Figure 6: A fully quantified fragment of the interval network associated with our semi-qualitative network.

## 7  Conclusions and further research

A first step in the quantification of a probabilistic network can be to elicit signs instead of numbers from a domain expert. We can then study the reasoning behaviour of the network under construction using the thus obtained qualitative probabilistic network. To bridge the gap between the coarse level of representation detail of a qualitative network and the level of detail of a quantified network, we have proposed to perform quantification in a stepwise manner, studying the reasoning behaviour of the resulting semi-qualitative network after each step. To support our methodology, we have introduced the formalism of semi-qualitative probabilistic networks. In addition, we have presented an algorithm for probabilistic inference in a semi-qualitative network that amounts to propagating intervals in its associated interval network.

The algorithm that we have presented for interval propagation becomes less efficient as more numerical information is added to the network under construction. This is, of course, not surprising given the computational complexity of inference in a probabilistic network in general. We have shown, however, that a polynomial bound can be put on the complexity if desired. We have further shown that the numerical information available in a semi-qualitative network can be exploited to tighten the intervals in its associated interval network. Further research is required to determine whether or not the available information can be exploited to an even further extent. We have shown that ambiguous intervals resulting from trade-offs in the network may be locally resolved using the methods provided by Liu and Wellman, as long as enough numerical information is available to apply these.

In conclusion, we feel that the stepwise methodology we have proposed provides for effective quantification of a probabilistic network. Each time a part of the network under construction is quantified, the reasoning behaviour of the resulting semi-qualitative network can be studied, thereby allowing for early identification of modelling inadequacies and for better understanding of the network by the domain experts. We feel that the robustness and quality of the network will ultimately benefit from the use of our methodology.

**Acknowledgements**

This research has been (partly) supported by the Netherlands Organisation for Scientific Research (NWO).